\documentclass[letterpaper,12pt]{article}   
\usepackage{osajnl2} 
\usepackage{hyperref}
\usepackage{amsmath}

\begin{document}

\title{No-reference image quality assessment through the von Mises distribution} 


\author{Salvador Gabarda$^{1}$ and Gabriel Crist\'{o}bal$^{1,*}$}
\address{$^1$Instituto de \'{O}ptica (CSIC), \\ Serrano 121, 28006 Madrid, Spain}
\address{$^*$Corresponding author: gabriel@optica.csic.es}

\begin{abstract}
An innovative way of calculating the von Mises distribution (VMD) of image entropy is introduced in this paper. The VMD's concentration parameter and some fitness parameter that will be later defined, have been analyzed in the experimental part for determining their suitability as a image quality assessment measure in some particular distortions such as Gaussian blur or additive Gaussian noise. To achieve such measure, the local R\'{e}nyi entropy is calculated in four equally spaced orientations and used to determine the parameters of the von Mises distribution of the image entropy. Considering contextual images, experimental results after applying this model show that the best-in-focus noise-free images are associated with the highest values for the von Mises distribution concentration parameter and the highest approximation of image data to the von Mises distribution model. Our defined von Misses fitness parameter experimentally appears also as a suitable no-reference image quality assessment indicator for no-contextual images. 

\end{abstract}


\maketitle 

\section{Introduction}
Digital images can suffer a number of different operations that take into account e.g. acquisition, coding, compression, transmission, and many others before their final use. In such processes, digital images might be affected by many kinds of degradations revealed by different types of visual distortions. A list of different possible types of degradations has been considered in the image database TID2008 by Ponomarenko et al. \cite{Ponomarenko2009}. A key feature about images is their quality. However, image quality has not a simple definition. In a semantic approximation, \emph{image quality is understood as the subjective impression of how well image content is rendered or reproduced} \cite{Yendrikhovskij2002}. This definition yields to the concept of image quality assessment. In practice, such assessment can be achieved by two means: one is by psychophysical subjective experiments with human observers, and the other one is by objective metrics applied directly to digital images \cite{Ciocca2009}. Image quality assessment is a very active research area with many contributions, techniques and models \cite{Wang2006}. Typically, image quality can be measured by comparison with a reference, but unfortnately such reference is not available in many applications.  Hence, no-reference image quality assessment methods are required and they have been the subject of a very active research during the last years. However, a universal method for estimating the overall image quality is still a challenging open issue. Customarily, existing no-reference methods deal with simple specific degradations such as Gaussian blur or additive white noise. A good survey about existing no-reference sharpness metrics may be found in Ferzli and Karan \cite{Ferzli2009} where their metric is compared to others with good results. Ferzli and Karan introduced the notion of \emph{just noticeable blur} (JNB). It is an edge-based sharpness metric based on a human visual system model. Other metrics have been developed including blur and noise simultaneously, such as the one by Zhu and Milanfar \cite{Zhu2010}, introducing a new concept called \emph{true image content}. Their measure is correlated with noise, sharpness, and intensity contrast, manifested in visually salient geometric features such as edges, showing that such measure correlates well with subjective quality evaluation for both blur and noise distortions. However, Zhu and Milanfar technique has been designed to compare images within the same context, while the sharpness metric of Ferzli and Karam has been developed to predict the relative amount of blurriness in images regardless of their context (note that we consider that images resulting from distorting a given original are contextual each other and we use the term no-contextual when the degraded images are the result of distorting different originals).  According to Zhu, JNB technique fails to capture the trend of quality change in BM3D \cite{Zhu2010} denoising experiments, since it cannot handle noise well. Later on, Narvekar and Karam \cite{Narvekar2010} proposed an improved algorithm based on the JNB paradigm for a no-reference objective image sharpness metric, introducing a technique they called the \emph{cumulative probability of blur detection} (CPBD). In such work, the sharpness metric converges to a finite number of quality classes. They used the LIVE \cite{LIVE2003} database to validate the performance of their metric. A training-based method determines the centroids of the quality classes that represent the perceived quality levels. Classification is based on assigning the image to one of the quality classes and then using the index of the corresponding quality class as the metric value for that image. They include measuring experiments for Gaussian blur and JPEG2000-compressed images and they show that this metric performs better than other known metrics.

In summary, different reference and no-reference classifications of metrics can be considered. According to the type of distortion they measure,  \emph{specific metrics} (only one type) versus \emph{universal metrics} (all types) could be considered. According to their contextual character,  \emph{contextual metrics}, if the measures have sense only when the images have the same context versus \emph{no-contextual metrics}, if the measure have sense by itself, regardless of the image context could be considered. In addition to the previous schemes, supervised and unsupervised approaches may be considered in the image quality context. 

This paper is a step forward to the approach introduced by Gabarda and Crist\'{o}bal \cite{Gabarda2007} that links image directional entropy with image quality assessment. In \cite{Gabarda2007}, the  variance of the directional entropy of the image was introduced as a quality assessment parameter, presenting the highest value when the image is free of Gaussian blur or Gaussian noise and decreases monotonously with blur an noise increments. Gabarda and Crist\'{o}bal's method may be classified as a no-reference, specific, contextual and unsupervised metric. In this paper we extend this methodology to both contextual and no-contextual applications for image quality assessment by introducing the von Mises distribution of the directional image entropy as the cornerstone of a new image quality metric for blur and noise.
   
This paper is organized as follows. Section 2 gives the mathematical background required to understand our methodology. Section 3 presents some applications of the VMD for image quality measurement, along with the performances of this method against the Mean Opinion Score (MOS) of TID2008 image database and Differential Mean Opinion Score (DMOS) of LIVE image database. Results are compared with scores from other no-reference quality assessment existing methods. Finally, conclusions are drawn in Section 4.

\section{Description of the method}
\subsection{Directional entropy}
The information content of images can be measured in different ways. One of the most popular is the entropy. Generally speaking, the measurement of entropy was initially proposed by Shannon \cite{Shannon1949} as a measure of the information content per symbol, coming from a stochastic information source. It can be considered as a quantifier of the uncertainty or randomness of a signal or, in other words, a measure of the information content of the signal.
Given a discrete random variable $X=\lbrace x_1,x_2,...,x_I \rbrace$, the Shannon entropy of this discrete random variable can be computed as
\begin{equation}
H(X)=-\sum\limits_{i=1}^I P(x_i) \log P(x_i)
\end{equation} 
where $P(x_i)$ is the \emph{probability} of event $x_i$.

Later, R\'{e}nyi \cite{Renyi1976} extended this notion to yield the generalized entropy, whose representation for discrete random variables is
\begin{equation}\label{eq:Renyi}
R_\alpha (X)=\frac{1}{1- \alpha} \log_2 \left( \sum\limits_{i=1}^I P^\alpha(x_i) \right)
\end{equation}
Here $ \alpha $ is a real-positive number. It can be shown that the R\'{e}nyi entropy, as a generalization of the Shannon entropy, reverts to the Shannon entropy when $ \alpha \rightarrow 1 $.

Measures of entropy may be global, affecting the whole image, or may be localized in each pixel by the use of an sliding window, centered in each pixel, that gathers the neighborhood of that pixel with the desired geometry.

 Our proposal is to use a linear array of pixel values, centered in the interest pixel and oriented in a set of desired directions, in order to have a directional measure of entropy at pixel level. To do so, an operational sliding window \emph{L}  is defined, containing the $ N+1 $ pixels that better fit the ideal line centered in the interest pixel  and directed by angle  $ \theta $. \emph{N} is required to be an even number to preserve the symmetry of the window.
  
Better than considering the gray-values of the image as input to the entropy calculation, we prefer considering the space-frequency information of the image. Space-frequency representations of a signal open up the possibility of exploring the image information taking into account the position and the local frequency content of the image.
One of these space-frequency (SF) representations is the Wigner distribution \cite{Wigner1932}. A common analytical framework of SF representations was introduced by Cohen \cite{Cohen1966}. The Wigner distribution (WD) has been selected due to its excellent properties and because it can be regarded as a {\it masterform} distribution function from which the other existing representation can be derived as filtered versions of it \cite{Jacobson1988}. The WD approximation for discrete signals is better known as pseudo-Wigner distribution (PWD). To calculate a local PWD we will use the following equation
\begin{equation}\label{eq:Wigner}
W(n,k)=2 \sum\limits_{m=- \frac{N}{2}}^{\frac{N}{2}-1} z(n+m)z^*(n-m) e^{-i2 \pi k \left(\frac{2m}{N} \right) }
\end{equation} 
                 
This approximation of the PWD is due to Claasen and Mecklembräuker \cite{Claasen1980} and is similar to Brenner's expression \cite{Brenner1983}. 
In Eq. (\ref{eq:Wigner}) the variable $z(n)$ represents the gray value of pixel $n$ in a given image $z$. A directional PWD may be calculated by using Eq. (\ref{eq:Wigner}) along with a pixel-by-pixel sliding window $D_{\theta}$  to pick up the values $z=z(-N/2), ..., z(0), ..., z(N/2)$. The central position of the window matches exactly pixel $n$ on the image. Eq. (\ref{eq:Wigner}) can be interpreted as the discrete Fourier transform (DFT) of the product  $r(n,m)= z(n+m)z^*(n-m)$. Here $z^*$ indicates the complex-conjugate of signal $z$ (note that $z=z^*$ for real valued images). The sum is limited to a spatial interval $(-N/2,N/2-1)$. In Eq. (\ref{eq:Wigner}), $n$ and $k$ represent the space and frequency discrete variables respectively, and $m$ is a shifting parameter, which is also discrete. By shifting the window to all possible positions over the image, the full pixelwise PWD of the image is produced.

In what follows, we will take the PWD in Eq. (\ref{eq:Wigner}) as the distribution to be used as input to Eq. (\ref{eq:Renyi}). Consequently, each spatial position $n$ will receive a certain value of entropy.
 
Diverse measures of entropy may be defined upon the kind of distribution and normalization used with the formulation due to R\'{e}nyi. Although the R\'{e}nyi measures in time-frequency distributions formally resemble the original entropies, they do not have the same properties, conclusions and results derived in classical information theory. The positivity, $P(n,k) \geq 0$   will not be always preserved, along with the unity energy condition  $\sum_k P(n,k)=1$ for a given pixel $n$. In order to reduce a distribution to the unity signal energy case, some kind of normalization should be applied. The normalization can be performed in various ways, leading to a variety of possible measurement definitions \cite{Sang1995}, \cite{Williams1991}, \cite{Flandrin1994}, \cite{Pitton1993}, \cite{Stankovic2001}.
  
The R\'{e}nyi entropy measure applied to a discrete space-frequency distribution, as the PWD denoted by $W(n,k)$, can be written as
\begin{equation}\label{eq:Renyi_2}
R(n)=-\frac{1}{2} \log_2 \left( \sum\limits_{k}W^3 (n,k) \right)
\end{equation}
                       
Here $\alpha=3$  and we have dropped the subscript $\alpha$. The summation is restricted to the 1-D neighborhood of pixel $n$, as described above. Again $k$ is the frequency discrete variable. In order to reduce the distribution $W(n,k)$ to the unity signal energy case, some kind of normalization must be done \cite{Sang1995}. To do so, the PWD given by Eq.(\ref{eq:Wigner}) is normalized by applying the following normalizing law, as a previous step to the R\'{e}nyi entropy measure:

\begin{equation}
\tilde{W}(n,k)=\frac{W(n,k)W^*(n,k)}{\sum\limits_{k} \left( W(n,k)W^*(n,k) \right) }=\frac{W^2(n,k)}{\sum\limits_{k}W^2(n,k)}
\end{equation} 
                         
Squaring of $W$ is justified due to the real character of the PWD, and the normalizing step affect exclusively to index $k$, when the window operation in Eq. (\ref{eq:Renyi_2}) is applied to pixel $n$; hence the condition $\sum\limits_{k}\tilde{W}(n,k)=1$  in such position is fulfilled.

It can be shown that the R\'{e}nyi entropy meet the constraint $0 \leq R_\alpha(n) \leq \log_2 N$ . Hence, the measure can be normalized to the interval $[0,1]$ by applying $R(n)=R(n)/\log_2 N$ in Eq. (\ref{eq:Renyi_2}).

The 1-D window used to measure entropy by this method may be oriented in any desired direction $\theta$,  as indicated above, and entropy is now denoted by $R(n,\theta)$. This local pixel-wise directional entropy can be constrained to a scalar value for the whole image by averaging all the $R(n,\theta)$ values by means of

\begin{equation}\label{eq:Renyi_3}
\overline{R}(\theta)=\langle R(n,\theta) \rangle = \frac{1}{M}\sum\limits_{n}R(n,\theta)
\end{equation} 	
	                
provided that $M$ is the number of pixels in the image.

\subsection{The von Mises distribution}
The von Mises (VM) distribution belongs to probability theory and is used to handle directional statistics for continuous probability distribution on a circular basis. It appears, in many respects, analogous to the normal distributions for a scalar variable.  It was proposed by von Mises \cite{von_Mises1918} to study deviations of atomic weights from integer values. This distribution has also been applied to diverse applications in many fields and has become an important tool in the statistical theory of directional data \cite{Abramowitz1964}.

The von Mises probability density function for the angular variable $\theta$  is given by:
\begin{equation}
f(\theta | \mu,\kappa)=\frac{1}{2 \pi I_0(\kappa)}e^{\kappa \cos (\theta-\mu)}
\end{equation}              
where $I_0(\kappa)$ is the modified Bessel function of order $0$, and $\theta$ is defined in the interval $[-\pi,\pi)$. The parameter $\kappa$ is responsible for how concentrated the distribution is around the mean direction $\mu$. Larger values of the concentration parameter $\kappa$ indicate that the distribution is more closely grouped around the mean direction. When $\kappa=0$, this distribution is equivalent to the uniform distribution. This fact happens when directions in the image are endowed with equal probability. The uniform circular distribution is, for example, a good model for Gaussian noise.

Different authors have dedicated some research to the link between images and the von Mises distribution. For instance, Vo et al. \cite{Vo2008} proposed a new statistical modeling of natural images in the wavelet transform domain. They claimed that the von Mises distribution fits accurately the behaviors of relative phases in the complex directional wavelet sub-band from different natural images, and introduced a new image feature based on the VM model for image texture retrieval applications. Palacios et al.  \cite{Palacios2007} presented what they called new tools for color image processing, based on the circularity of the hue variable of a color image. They give a definition of the median and the range of angular data and apply their results to detect hue edges. Feng \cite{Feng2003} presented a local image orientation estimation method using the image gradient, using a combination of the principal component analysis (PCA) and the multiscale pyramid decomposition. Grana et al. \cite{Grana2008} described a new approach to texture characterization for document analysis. By considering the autocorrelation matrix, they described image texture through a mixture of VM distributions.
 
\subsection{Von Mises distribution of image entropy}
As a further contribution to the applications cited in the previous section, we propose to use the VM distribution for modeling the directional distribution of the image entropy, calculated by means of Eq. (\ref{eq:Renyi_3}). The measure defined for the image by Eq. (\ref{eq:Renyi_3}) requires the use of a bimodal von Mises distribution, provided that $\overline{R}(\theta)=\overline{R}(\theta+\pi)$. This requirement is fulfilled when we take $a=1/2$, and $\mu_2=\mu_1+\pi$ in the following bimodal expression of the von Mises distribution
\begin{equation}
f(\theta | \mu,\kappa)=\frac{a}{2 \pi I_0(\kappa)}e^{\kappa \cos (\theta-\mu_1)}+\frac{1-a}{2 \pi I_0(\kappa)}e^{\kappa \cos (\theta-\mu_2)}
\end{equation}  
from which we arrive to \cite{Stephens1969}
\begin{equation}\label{eq:VM_Stephens}
f(\theta | \mu,\kappa)=\frac{1}{2 \pi I_0(\kappa)}\cosh \left(\kappa \cos(\theta-\mu) \right)
\end{equation}

\begin{figure}[p] 
\centerline{\includegraphics[width=0.7\textwidth]{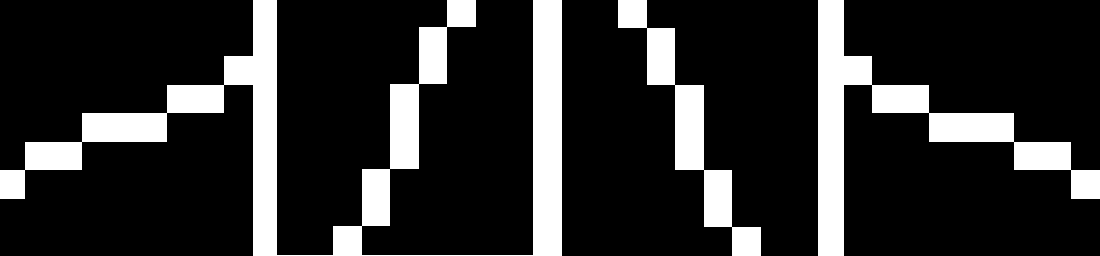}} 
\caption{Different configuration of pixels to define directional windows $D_\theta$, with $N=9$ pixels and four equally spaced orientations $\pi/8,3\pi/8,5\pi/8,7\pi/8$.  Note that periodicity is $\pi$, provided that $\theta$ and $\theta+\pi$ represent the same direction.}
\label{fig:window}
\end{figure}

Eq. (\ref{eq:VM_Stephens}) can be considered as the basis for modeling directional statistics of image entropy. One of the features to take into account in the discrete directional model for image entropy starting from Eq. (\ref{eq:Renyi_3}) concerns the number of axis that can be defined in the image. Two requirements must to be fulfilled in order to consider that the directional measures, based on Eq. (\ref{eq:Renyi_3}) are completely comparable. First all directions must be measured by windows consisting in the same number of pixels, and the second one is that the span of the windows in all possible directions must have the same length. These two constrains are only possible when using four directions along the axes of a regular octagon, i.e.: $\pi/8,3\pi/8,5\pi/8,7\pi/8$. Under this requirements, we are going to restrict our calculations upon 1D windows directed by the axes along these angles. The arrangement of the pixels in these four operational windows may be visualized in Fig. \ref{fig:window}.

The four directionalities $\theta_i \in \{\pi/8,3\pi/8,5\pi/8,7\pi/8\}$ that we have defined are used to calculate four values  $\overline{R}(\theta_i)$ of entropy by means of Eq. (\ref{eq:Renyi_3}) that it can be referred as vectors $\mathbf{R}_i=(R_i,\theta_i)$. These directional entropies can be represented also by means of Cartesian vectors by $\mathbf{R}_i=(x_i,y_i)^T=(R_i \cos \theta_i,R_i \sin \theta_i)^T$.
 
Some definitions for a resultant vector and a mean angular direction in VM distribution have been given by Jammalamadaka and SenGupta \cite{Jammalamadaka2001} and one analysis for modeling circular data may be found in Bentley \cite{Bentley2006}.

In our approximation we first perform a preliminary estimation of $\mu$ by a  single value decomposition (SVD) of the directional vectors $\mathbf{x}_i$, following to Feng \cite{Feng2003}, that deals the image orientation estimation problem through the local image gradient. The singular vector $\mathbf{v}_i$ of the matrix
\begin{equation}
\mathbf{X}=
\begin{pmatrix} 
\mathbf{R}_{1}^T   \\
\mathbf{R}_{2}^T   \\
\mathbf{R}_{3}^T   \\
\mathbf{R}_{4}^T 
\end{pmatrix}=
\begin{pmatrix} 
R_1 \cos \theta_1 & R_1 \sin \theta_1  \\
R_2 \cos \theta_2 & R_2 \sin \theta_2 \\
R_3 \cos \theta_3 & R_3 \sin \theta_3 \\
R_4 \cos \theta_4 & R_4 \sin \theta_4 
\end{pmatrix}
\end{equation}         
corresponding to its largest singular value, gives an estimation of the parameter $\hat{\mu}=\arg(\mathbf{v}_1)$.

Also, an initial estimation of $\kappa$ can be achieved by the following approximation due to Dhillon \cite{Dhillon2003}: $\hat{\kappa}=1/2(1-\overline{R})$, where $\overline{R}=\| \sum_{i=1}^4 \mathbf{R}_i \| /4$.

The second step is a gradient descent algorithm to find a better estimation for $\kappa$. The set of values $\{f(\theta_i),R_i\}$, for $\theta_i = \pi/8,3\pi/8,5\pi/8,7\pi/8$, with  $i=1, 2, 3, 4$ respectively, are matched by a minimum square error estimation method to find the coefficients $A$ and $B$ that fit the equation $R_i=Af(\theta_i)+B$. An exact fitting is attained if $A=1$ and $B=0$. An error function $\varepsilon(\kappa)=\|\left(A(\kappa),B(\kappa) \right)-(1,0)\|$  is defined to control the $\kappa$ increments, in order to find a minimum for this error function. The algorithm runs by means of a recursive law $\kappa_{i+1}=(1 \pm C) \times \kappa_i$, $C=0.01$ for updating the values of $\kappa$. The algorithm stops when the error value reaches a minimum. A Matlab implementation of this algorithm may be found in \cite{Matlab2012}.

\begin{figure}[p]
\centerline{\includegraphics[width=0.7\textwidth]{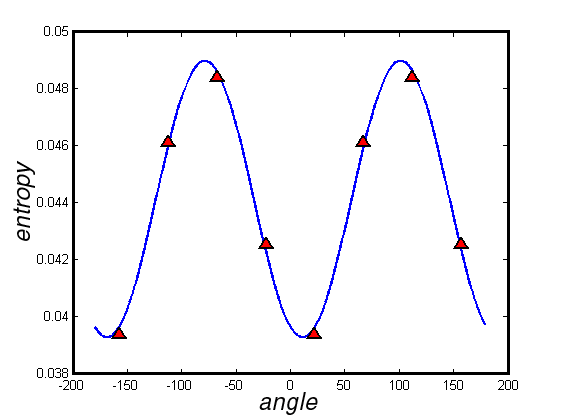}} 
\caption{Von Mises distribution of the directional entropy for image I21 contained in the database TID2008 \cite{Ponomarenko2009}. Triangles indicate the value of directional entropy for the target image after four different axes. $\kappa=-0.36$, $\mu=-78.75$ deg (twin pick at $101.25$ deg.). Fitness parameter : $\varphi=0.89$.}
\label{fig:von_Mises}
\end{figure}

The accuracy of the estimated distribution $\hat{f}(\theta)$ to a true von Mises distribution is measured by defining a fitness parameter $\varphi=e^{-\varepsilon}$, whose value will be $1$ for the best fitting and $0$ for the worst. An example of estimation of the von Mises distribution for a given image is shown in Fig. \ref{fig:von_Mises}. Triangles indicate the directional entropy values $(R_i,\theta_i)$ used as input to the algorithm, along with the estimated von Mises distribution, $\hat{f}(\theta)$, of the image.

To characterize the behavior of $\kappa$ and $\varphi$, the algorithm has been tested with the images in the image database TID2008 due to Ponomarenko et al. \cite{Ponomarenko2009}. This database contains 25 reference color images and 1700 distorted images (25 reference images $\times$ 17 types of distortion $\times$ 4 levels of distortion) in bitmap format without any compression. Images are $512 \times 284$ pixels in size. The test  has been performed taking the 25 reference images, excluding image 25th, provided that this last one is artificial and has been considered inappropriate for our study.

\begin{figure}[p]
\centerline{\includegraphics[width=0.7\textwidth]{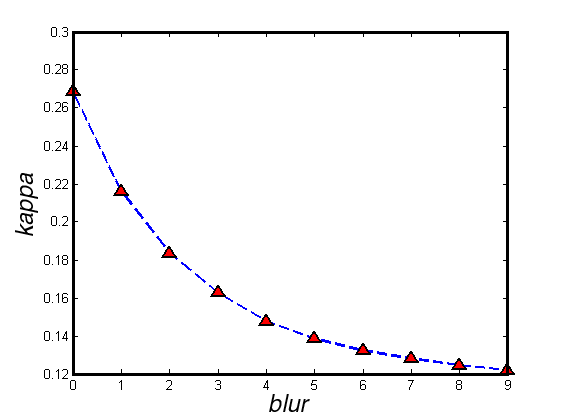}} 
\caption{Averaged variation of $\kappa$ against increasing amount of Gaussian blur for reference images in TID2008 database.} 
\label{fig:kappa_blur_TID2008}
\end{figure}

\begin{figure}[p]
\centerline{\includegraphics[width=0.7\textwidth]{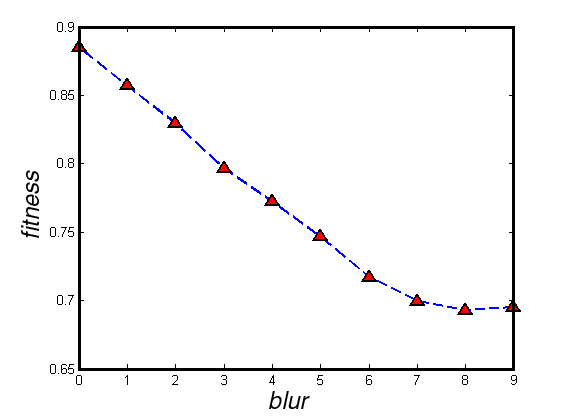}} 
\caption{Averaged variation of $\varphi$ (fitness) against increasing amount of Gaussian blur for reference images in TID2008 database.}
\label{fig:phi_blur_TID2008}
\end{figure}

Reference images have been first degraded  by adding blur in iterative manner by means of a rotationally symmetric Gaussian lowpass filter of size $5 \times 5$ pixels with standard deviation $\sigma=1$, and in a second instance by adding Gaussian noise, also iteratively, with standard deviation $\sigma=0.01$. Considering the 24 originals and nine iterations for each original image, we produced a test set of 240 images. Figures \ref{fig:kappa_blur_TID2008} and \ref{fig:phi_blur_TID2008} show the average values of $\kappa$ and $\varphi$ for all these images, having the degradation level as abscissa. Measurements indicated that $\varphi=0.88 \pm 0.02$ as the expected value of the fitness parameter for the 24 originals (see abscissa=0 in Fig. \ref{fig:phi_blur_TID2008}), indicating that the fitness parameter $\varphi$ of the VM distribution converges to a common value for good quality images. In general, when more blur or Gaussian noise is added, fitness decreases, and the variance of   $\varphi$ increases, indicating that, for severe degradation, the von Mises entropy paradigm for image modeling does not hold any more.

\begin{figure}[p]
\centerline{\includegraphics[width=0.7\textwidth]{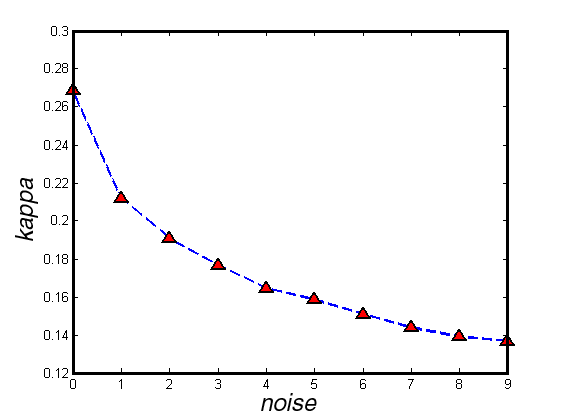}} 
\caption{Averaged variation of $\kappa$ against increasing amount of Gaussian noise for reference images in TID2008 database.}
\label{fig:kappa_noise_TID2008}
\end{figure}

\begin{figure}[p]
\centerline{\includegraphics[width=0.7\textwidth]{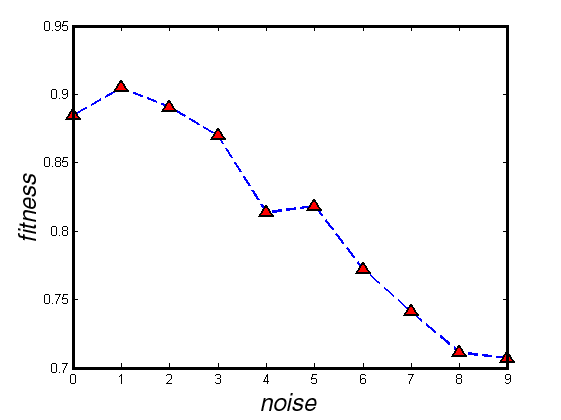}} 
\caption{Averaged variation of $\varphi$ (fitness) against increasing amount of Gaussian noise for reference images in TID2008 
database.}
\label{fig:phi_noise_TID2008}
\end{figure}

\begin{figure}[p]
\centerline{\includegraphics[width=0.7\textwidth]{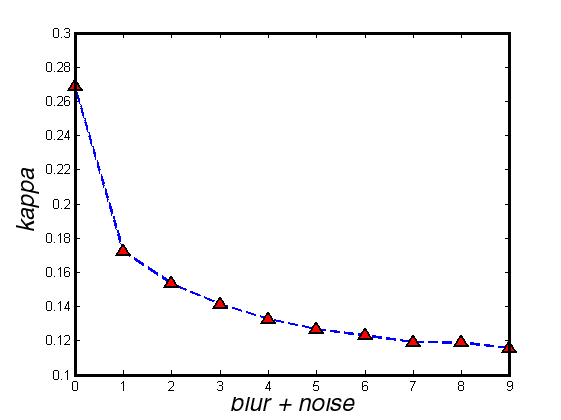}} 
\caption{Averaged variation of $\kappa$ against increasing amount of Gaussian blur and a fixed amount of Gaussian noise for reference images in TID2008 database.}
\label{fig:kappa_blurandnoise_TID2008}
\end{figure}

\begin{figure}[p]
\centerline{\includegraphics[width=0.7\textwidth]{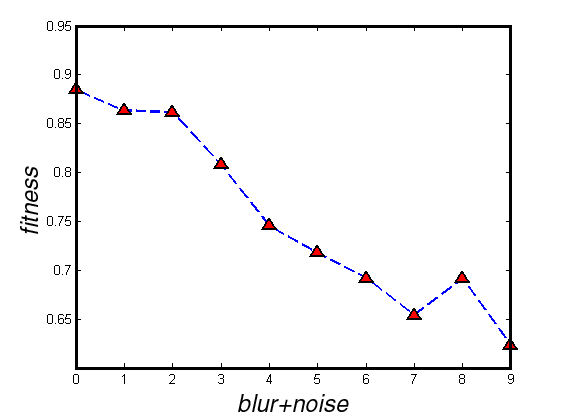}} 
\caption{Averaged variation of  $\varphi$ (fitness) against increasing amount of Gaussian blur and a fixed amount of Gaussian noise for reference images in TID2008 database.}
\label{fig:phi_blurandnoise_TID2008}
\end{figure}

Figures \ref{fig:kappa_blur_TID2008} to \ref{fig:phi_noise_TID2008} show the behavior of the parameters $\kappa$ and    $\varphi$ by averaging the values from the outcome of our algorithm for the 24 images in TID2008 database. A hybrid blur-noise situation has been experimented by adding blur and noise simultaneously to the images in TID2008 database (see Fig. \ref{fig:kappa_blurandnoise_TID2008} and \ref{fig:phi_blurandnoise_TID2008}.
 
Two main consequences are derived from the previous observations. First, the fitness parameter has a low value ($\varphi<<0.88$) for blurred images or very noisy images, and a high value ($\varphi \cong 0.88$) for undegraded images. The second consequence is that quality of the images is maximum when $\kappa$ reaches a maximum in the series of diversely degraded images from a given original.

\section{Application of VM distribution to image quality assessment}

\subsection{VM distribution and image quality assessment of contextual images}
The behavior of the $\kappa$ parameter of the von Mises distribution suggest that it can be considered as an image quality assessmente index, when dealing with Gaussian blur or Gausssian noise. In order to have a quantitative evaluation of this statement, we compare the classification after this  parameter vs. the Mean Opinion Score (MOS) for two specified distortion types in database TID2008.

The image database TID2008 includes 17 different degradation types with one to four strength levels. The image database includes information about the MOS for each distorted image. The degraded images labeled as Gaussian noise, and Gaussian blur have been scored by means the  $\kappa$ value  determined by our VM algorithm. Later on, Kendall, Spearman and Pearson correlation coefficients have been evaluated by comparing our scores with the MOS of the images in TID2008 image database.  Results are shown in Table \ref{tab:Table_1}. Figures in Table \ref{tab:Table_1} must be understood as  measures in a contextual, specialized, no-reference and no-learning metric.

In order to put figures in Table \ref{tab:Table_1} in line with other existing no-reference methods, we include Table \ref{tab:Table_2}, where our measures are compared to the results published by Zhu and Milanfar \cite{Zhu2010} using the same image database and the Spearman correlation coefficient.

Figures in Table \ref{tab:Table_2} reveal that the $\kappa$ parameter of VM distribution has a strong correlation with Zhu and Milanfar measures   for Gaussian  blur and Gaussian noise distortions.

\subsection{Autofocusing example}
A short depth of field is an inherent limitation in optical microscopy. Basically, the focal plane of the microscope needs to be critically located to observe the sharper image. Hence, auto-focusing techniques must be included in automatized optical microscope systems. Relevant examples of images from bright field microscopy may be found in Valdecasas \cite{Valdecasas2001}. 

\begin{figure}[p]
\centerline{\includegraphics[width=0.7\textwidth]{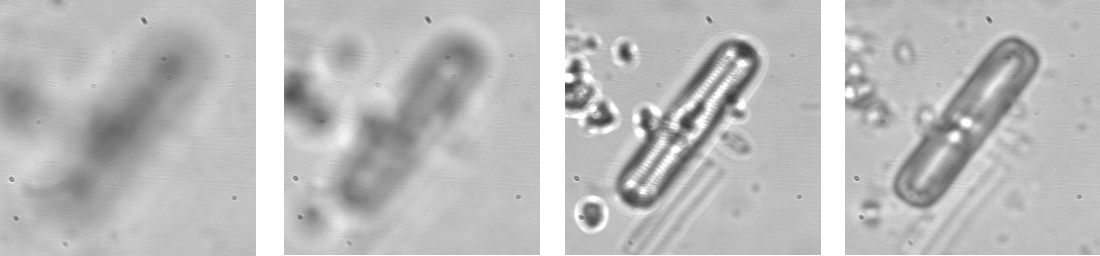}} 
\caption{From left to right, capture 10, 30, 50, and 70 from a sequence of 100 light microscope images.}
\label{fig:test_images}
\end{figure}

\begin{figure}[p]
\centerline{\includegraphics[width=0.7\textwidth]{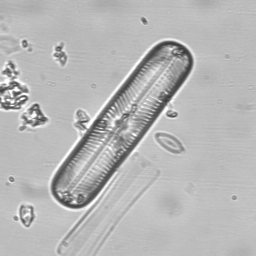}} 
\caption{The best quality image in the sequence (no. 57) after the von Mises algorithm.}
\label{fig:best_image}
\end{figure}

\begin{figure}[p]
\centerline{\includegraphics[width=0.5\textwidth]{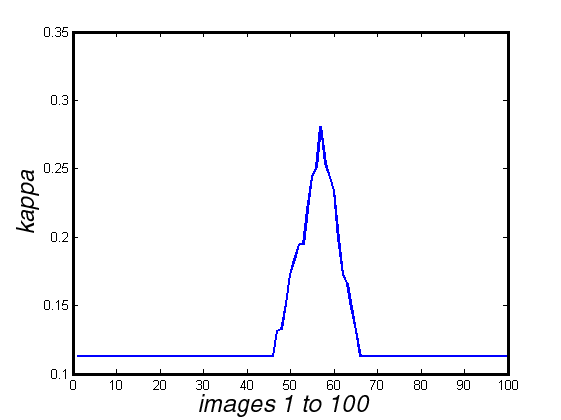}} 
\caption{$\kappa$ values for the whole sequence of microscopic images. The maximum of $\kappa$ correspond to image 57}
\label{fig:test_graphic}
\end{figure}

The behavior of the $\kappa$ parameter of the von Mises distribution suggest that it can be considered as an image quality assessmente index, when dealing with Gaussian blur or Gausssian noise. After the above quantitative evaluation of this parameter through the TID2008 database, we present a real life example to illustrate one of the possible applications of this no-reference image quality assessment method for contextual images. In this example, a sequence of 100 images from an optical microscope have been processed by our algorithm to determine their $\kappa$ values. Images are $256 \times 256$ pixels in size. Images 10, 30, 50 and 70 of the stack are shown in Fig. \ref{fig:test_images}. The best image according to our algorithm is number 57 in the sequence, that can be seen in Fig. \ref{fig:best_image}. Fig. \ref{fig:test_graphic} plots the  values of $\kappa$ for the whole sequence of images.

\subsection{VM distribution and image quality assessment of no-contextual images}

One of the objectives of this paper consists in determining whether the VM distribution can be used to construct a no-reference quality metric in the no-contextual case. The definition of a metric for  image quality assessment based on the VM distribution for no-contextual images requires some kind of normalization, to assure independence of the measure from the image context. To do so, we have designed two possible measures taking into account the results in the previous section. Following these results, we postulate that both, the concentration parameter, $\kappa$, and the fitness parameter, $\varphi$, decrease exponentially with increasing amount of degradation (e.g.: Gaussian blur), that can be modelled by expressions like $\kappa=\kappa_0 \exp(-\beta D)$ or  $\varphi=\varphi_0 \exp(-\beta D)$, where $D$ is the degradation and  $\beta$ is a shape factor exclusive for each image. Empirically, we have found convergence at the origin for $\varphi_0$  to 0.88, although there is not a similar convergence for $\kappa_0$. The second of these expressions is simpler to calculate, provided that $\kappa_0$  is an unknown parameter, while $\varphi_0$ has been experimentally determined as $\varphi_0=0.88 \pm 0.02$ (for Gaussian blur). The exponential variation of $\varphi$ can be written in a logarithmic way as
\begin{equation}\label{eq:law}
\log \varphi=\log \varphi_0-\beta D
\end{equation}
The derivative of Eq. (\ref{eq:law}) gives  $d(\log \varphi) / dD = -\beta$, and $\beta$ can be approximated by
\begin{equation}\label{eq:beta} 
\beta=-\Delta (\log \varphi ) / \Delta D 
\end{equation}
for small increments, $\Delta D$.
 
Now, let us suppose that an arbitrary image $I_i$ produces an outcome parameter $\varphi_i$. By convolving the image with an appropriate kernel, $g$, (i.e.: a PSF of $5 \times 5$ pixels and $\sigma=1.5$) a controlled amount of blur may be added to the image. Then, $I_{i+1}=I_i \ast g$ is a little more degraded image than $I_i$. Supposing that the fitness parameter for $I_{i+1}$  is $\varphi_{i+1}$, and making $\Delta D=1$ for a given measuring scale, we can say, from Eq. (\ref{eq:beta}), that $\beta \simeq \log(\varphi_i / \varphi_{i+1})$ and introducing this value in (\ref{eq:law})
\begin{equation}\label{eq:degradation}
D_i=-(\log \varphi_i - \log \varphi_0)/\beta
\end{equation}
Eq. (\ref{eq:degradation}) defines in this way a new no-contextual measure for image quality assessment for the Gaussian blur case. The measure requires the estimation of the decay constant $\beta$ for each specific image as indicated in Eq. (\ref{eq:beta}). This new measure will be referred here after as \emph{von Mises degradation measure} (VMDM) and $D$ will be expressed in VM degradation units.

In order to evaluate the performance of the VMDM, we have applied our algorithm to the Gaussian blur images in the LIVE image database. This group is composed by 144 degraded images plus 29 originals. For our validation test, we have selected a group of 29 images (20 percent of the degraded subgroup), i.e.: from image number 117 to 145. Then we have correlated our quality scores with the DMOS scores of this group. Considering that the observations are positive quantities varying over many orders of magnitude, it is plausible to assume that the noise will be Gaussian, and the data will be well modeled as a Gaussian process. In this context, it is standard practice in the statistics literature to take the log of the data \cite{Snelson2004}. Then, we have used $\log(1+D_i)$ instead of $D_i$ for correlation. Results for the Pearson and Spearman coefficients are shown in Table \ref{tab:Table_3}. 

As expected, Table \ref{tab:Table_3} presents good peformance results, but still far from a perfect matching with the DMOS scores. An objective measure must necessarily have some differences when compared with recordings from subjective human observers. Even more, DMOS values will change if the database is created using different methodologies and environmental settings; hence, subjective scores are not comparable \cite{Redi_2010}, while the objective test will remain the same. However, the scores in Table \ref{tab:Table_3} can be improved by applying some weighting transformation to the VMDM scores by incorporating to these scores some learning process that mimic the specific human preferences existing in the DMOS recordings accompanying the LIVE database. To do so, the logarithmic VMDM measures and the transformed VMDM measures may be related, for example, through a function $f$ \cite{Snelson2004} given by
\begin{eqnarray}
\hat{D}=f(D;\Psi)= D+\sum_{i=1}^{I} a_i \tanh \left (b_i(D+c_i) \right ) & a_i,b_i \geq 0 & \forall i
\end{eqnarray}
where $\Psi=\{a,b,c \}$.  

Taking $I=5$, the values for the $a_i,b_i,c_i$ coefficients have been determined by a Monte Carlo process that maximizes the Pearson+Spearman correlation with the DMOS scores. We have used images 1 to 116 (the 80 percent of the 145 blurred images) in the LIVE database as the learning set and we have used the former set of 29 images for crossvalidating the measures. Our results are shown in Table \ref{tab:Table_4}. The table includes also the results from Narvekar and Karam's method for the same group of distortion. 

Comparative results indicate that this method based on the $\varphi$ parameter of the VM distribution is highly matching the DMOS scores for the Gaussian blur set of images in the LIVE database. Our results are comparable with the results given by Narvekar and Karam's method which is based on a sharpness metric for Gaussian blurring when applied to no-contextual images. Our method has the advantage to be self-contained and its performance can be even improved by adding a learning step.

\section{Conclusions}
In this paper, we have introduced a new way of determining the VM distribution of the image information. The possible applications of the VM distribution for image quality assessment have been experimentally tested. The $\kappa$ parameter of the VM distribution has experimentally shown that can be considered as a suitable no-reference quality indicator when dealing with contextual images. Significative results have been presented for the Gaussian blur and Gaussian noise cases. Also, the defined VM distribution fitness parameter has shown its suitability as a no-reference quality indicator in the no-contextual case for blur images. At the same time as the relationship of the VM distribution to image quality assessment has been shown in this paper, this new methodology suggest new applications of the VM distribution of image entropy for image processing, that will be the subject of future work.

\clearpage

\clearpage

\clearpage

\clearpage

\clearpage

\clearpage

\clearpage

\clearpage

\clearpage

\clearpage

\clearpage

\clearpage

\begin{table}[p]
\caption{VMDM Correlation Coefficients}
\centering 
\begin{tabular}{c c c c} \hline
Type of noise & Kendall & Pearson & Spearman \\ \hline
Gaussian noise & 0.7778  & 0.8052 & 0.8083   \\
Gaussian blur & 1.0000  & 0.9600 & 1.0000 \\
 \hline
\end{tabular}
\label{tab:Table_1}
\end{table}

\clearpage

\begin{table}[p]
\caption{Spearman Correlation Coefficients}
\centering 
\begin{tabular}{c c c c} \hline
Method & Gaussian blur & Gaussian noise & Blur + noise \\ \hline
Zhu and Milanfar & 1.0000  & 0.9760 & 0.9210   \\
VMDM & 1.0000  & 0.8083 & 0.8083 \\
 \hline
\end{tabular}
\label{tab:Table_2}
\end{table}

\clearpage

\begin{table}[p]
\caption{Correlation VMDM-DMOS in LIVE database for Gaussian blur }
\centering 
\begin{tabular}{c c} \hline
Pearson & Spearman \\ \hline
0.8546  & 0.9275 \\
\hline
\end{tabular}
\label{tab:Table_3}
\end{table}

\clearpage

\begin{table}[p]
\caption{Comparison of correlation for two methods in LIVE database for Gaussian blur}
\centering 
\begin{tabular}{c c c} \hline
Method & Pearson & Spearman \\ \hline
Narvekar and Karam & 0.9211 & 0.9449 \\ 
transformed VMDM & 0.8926 & 0.9311 \\
\hline
\end{tabular}
\label{tab:Table_4}
\end{table}


\begin{thebibliography}{99}

\bibitem{Ponomarenko2009}
	N. Ponomarenko, V. Lukin, A. Zelensky, K. Egiazarian, M. Carli, F. Battisti, 
  ``A Database for Evaluation of Full-Reference Visual Quality Assessment Metrics'',
  Advances of Modern Radioelectronics, Vol. 10, pp. 30-45, 
  (2009).
  
 \bibitem{Yendrikhovskij2002}
  S. Yendrikhovskij,
  \emph{Image quality and colour categorisation}, 
  In: Lindsay, W., Mac-Donald, M. \& Ronnier, L. (eds.) Colour image science: exploiting digital media. Chichester, England,
  John Wiley \& Sons Ltd. pp. 393-420.
  (2002).
  
 \bibitem{Ciocca2009} 
  G. Ciocca, F. Marini, R. Schettini,
  ``Image quality  assessment in multimedia applications'',
   Proc. SPIE Electronic Imaging Conference on Multimedia Content Access: Algorithms and Systems III, Vol. 7255, 72550A,
   San Jose, 
   (2009).
 
 \bibitem{Wang2006}
 Z. Wang, A. Bovik, ``Modern image quality assessment'', Synthesis Lectures on Image, Video, and Multimedia Processing 2, pp. 1-156,
 (2006). 
 
 \bibitem{Ferzli2009}
 R. Ferzli, L. J. Karam,  ``A no-reference objective image sharpness metric based on the notion of just noticeable blur (JNB)'',
 IEEE Trans Image Process 18, pp. 717-728,
 (2009). 
 
 \bibitem{Zhu2010}
 X. Zhu, P. Milanfar, ``Automatic parameter selection for denoising algorithms using a no-reference measure of image content'', IEEE Trans Image Process 19, pp. 3116-3132,
 (2010).
  
 \bibitem{Narvekar2010}
 N. D. Narvekar and L. J. Karam,
 ``An Improved No-Reference Sharpness Metric Based on the Probability of Blur Detection'',
  International Workshop on Video Processing and Quality Metrics for Consumer Electronics (VPQM),
   http://www.vpqm.org [pdf] [Software],
   (2010).
   
 \bibitem{Gabarda2007}
 S. Gabarda and G. Crist\'{o}bal, ``Blind image quality assessment through anisotropy'',
  J Opt Soc Am A Opt Image Sci Vis 24, pp. B42-51
  (2007)
  
  \bibitem{Shannon1949}
  C. E. Shannon and W. Weaver,
  ``The Mathematical Theory of Communication'',
   The University of Illinois Press, Urbana, Chicago, London, 
   (1949)
   
 \bibitem{Renyi1976}
 A. R\'{e}nyi,
 \emph{Some fundamental questions of information theory},
  In Pál Turán, editor, Selected Papers of Alfred R\'enyi, volume 3, pp. 526-552. Akad\'emiai Kiad\'o, Budapest,
  (Originally: MTA III. Oszt. K$\ddot{o}$zl., 10, 1960, pp. 251-282), 
  (1976).
  
  \bibitem{Wigner1932}
  E. Wigner,
   ``On the quantum correction for thermodynamic equilibrium'',
    Phys. Rev. 40, pp. 749-759,
    (1932).
    
 \bibitem{Cohen1966}
 L. Cohen,
  ``Generalized phase-space distribution functions'',
   J. Math. Physics, Vol. 7, pp. 781-786,
   (1966).
   
 \bibitem{Jacobson1988}
 L. D. Jacobson and H. Wechsler,
 ``Joint spatial/spatial-frequency representation'',
  Signal processing, Vol. 14, pp. 37-68,
  (1988).

\bibitem{Claasen1980}
T. A. C. M. Claasen and W. F. G. Mecklenbräuker,
``The Wigner distribution - A Tool for Time Frequency Analysis'',
 Parts I-III." Philips J. Research, Vol. 35, pp. 217-250, pp. 276-300, pp. 372-389,
 (1980).
 
 \bibitem{Brenner1983}
 K. H. Brenner,
 ``A discrete version of the Wigner distribution function'',
  Proc. EURASIP, Signal Processing II: Theories and Applications, pp. 307-309,
  (1983).
 
 \bibitem{Sang1995}
 T.H. Sang, W.J. Williams,
 ``R\'enyi information and signal dependent optimal kernel design'',
 Proceedings of the ICASSP, vol. 2, pp. 997-1000,
 (1995).
 
 \bibitem{Williams1991}
W.J. Williams, M.L. Brown, A.O. Hero,
``Uncertainty, information and time-frequency distributions'',
SPIE Adv. Signal Process. Algebra Arch. Imp. 1566, pp. 144-156,
(1991).

\bibitem{Flandrin1994}
P. Flandrin, R.G. Baraniuk, O. Michel,
``Time-frequency complexity and information'',
Proceedings of the ICASSP, vol. 3, pp. 329-332,
(1994).

\bibitem{Pitton1993}
J. Pitton, P. Loughlin and L. Atlas,
``Positive time-frequency distributions via maximum entropy deconvolution of the evolutionary spectrum'',
Proc. ICASSP, vol. IV, pp. 436-439,
(1993).

\bibitem{Stankovic2001}
L. Stankovic,
``A measure of some time-frequency distributions concentration'',
Signal Processing, 81, 621-631,
(2001).

\bibitem{von_Mises1918}
R. von Mises,
Uber die ``Ganzzahligkeit'' der Atomgewicht und verwandte Fragen,
Physikalische Z., 19, 490-500,
(1918).

\bibitem{Abramowitz1964}
M. Abramowitz and I. A. Stegun,
\emph{Handbook of Mathematical Functions},
National Bureau of Standards, (reprinted Dover Publications, 1965. ISBN 0-486-61272-4),
(1964).

\bibitem{Stephens1969}
M. A. Stephens,
``Techniques for directional data'',
Technical report no. 150, Department of Statistics, Stanford University, California,
(1969).

\bibitem{Vo2008}
A.P.N. Vo, S. Oraintara,
``Statistical Image Modeling Using von Mises Distribution in the Complex Directional Wavelet Domain'',
IEEE International Symposium on Circuits and Systems, pp. 2885-2888
(2008).

\bibitem{Palacios2007}
A. R. Palacios, C. Rodr\'{\i}guez, C. Vejarano,
``Circular processing of the hue variable a particular trait of colour image processing'',
VISAPP, pp. 69-78,
(2007).

\bibitem{Feng2003}
X. Feng,
``The analysis and approaches to image local orientation estimation'',
Master Thesis, University of California, Santa Cruz,
(2003).

\bibitem{Grana2008}
C. Grana, D. Borghesani, R. Cucchiara,
``Describing Texture Directions with Von Mises Distributions'',
Proceedings of ICPR, pp.1-4,
(2008).

\bibitem{Jammalamadaka2001}
S. R.Jammalamadaka and A. SenGupta,
\emph{Topics in Circular Statistics},
World Scientific: New Jersey,
(2001).

\bibitem{Bentley2006}
J. Bentley,
\emph{Modelling circular data using a mixture of Von Mises and uniform distributions},
Department of Statistics and Actuarial Science - Simon Fraser University,
(2006).

\bibitem{Dhillon2003}
I. S. Dhillon and S. Sra,
\emph{Modeling Data using Directional Distributions},
Technical Report \# TR-03-06, Department of Computer Sciences, The University of Texas at Austin,
(2003).

\bibitem{Matlab2012}
http://www.mathworks.com/matlabcentral/fileexchange/authors/127745.

\bibitem{Valdecasas2001}
A.G. Valdecasas, D. Marshall, J.M. Becerra and J.J. Terrero,
``On the extended depth of focus algorithms for bright field microscopy'',
Micron, 32, pp. 559-569,
(2001).

\bibitem{LIVE2003}
H. R. Sheikh, A. C. Bovik, L. Cormack and Z. Wang,
``LIVE Image Quality Assessment Database'',
http://live.ece.utexas.edu/research/quality,
(2003).

\bibitem{Snelson2004}
E. Snelson, C. E. Rasmussen, Z. Ghahramani,
\emph{Warped Gaussian Processes, Advances in Neural Information Processing Systems},
16, edited by S. Thrun, L. Saul, and B. Schlkopf,
(2004).

\bibitem{Redi_2010}
J. Redi, R. Zunino, H. Liu, H. Alers,  I. Heynderickx,
``Comparing subjective image quality measurement methods for the creation of public databases'',
Proc. SPIE 7529, 752903 doi:10.1117/12.839195,
(2010).
 
\end{thebibliography}
\end{document}